\documentclass[11pt,a4paper,twocolumn]{article}
\usepackage{amsmath}
\usepackage{amssymb}
\usepackage{amsfonts}
\usepackage{graphicx}
\usepackage{color}
\usepackage{subcaption}

\newcommand{\mathbbR}{\mathbb{R}}

\newcommand{\boldone}{{\boldsymbol{1}}}

\newcommand{\boldH}{{\mathbf{H}}}
\newcommand{\boldI}{\mathbf{I}}

\newcommand{\boldK}{{\mathbf{K}}}
\newcommand{\boldL}{{\mathbf{L}}}

\newcommand{\boldX}{\mathbf{X}}

\newcommand{\boldx}{\mathbf{x}}
\newcommand{\boldy}{\mathbf{y}}

\newcommand{\boldalpha}{{\boldsymbol{\alpha}}}

\newcommand{\calD}{{\mathcal{D}}}

\date{}

\begin{document}

\title{Structural Explanations for Graph Neural Networks using HSIC}

\author{
Ayato Toyokuni${}^{\,1,2}$
\quad\hspace{-5pt}
Makoto Yamada${}^{\,1,2,3}$
\\
${}^{1}$ Kyoto University \quad
${}^{2}$ RIKEN AIP\quad
${}^{2}$ OIST\\
{\small \tt toyokuni.ayato@ml.ist.i.kyoto-u.ac.jp, makoto.yamada@oist.jp,}
}

\maketitle

\begin{abstract}
Graph neural networks (GNNs) are a type of neural model that tackle graphical tasks in an end-to-end manner. Recently, GNNs have been receiving increased attention in machine learning and data mining communities because of the higher performance they achieve in various tasks, including graph classification, link prediction, and recommendation. However, the complicated dynamics of GNNs make it difficult to understand which parts of the graph features contribute more strongly to the predictions. To handle the interpretability issues, recently, various GNN explanation methods have been proposed. In this study, a flexible model agnostic explanation method is proposed to detect significant structures in graphs using the Hilbert-Schmidt independence criterion (HSIC), which captures the nonlinear dependency between two variables through kernels. More specifically, we extend the GraphLIME method for node explanation with a group lasso and a fused lasso-based node explanation method. The group and fused regularization with GraphLIME enables the interpretation of GNNs in substructure units. Then, we show that the proposed approach can be used for the explanation of sequential graph classification tasks. Through experiments, it is demonstrated that our method can identify crucial structures in a target graph in various settings.
\end{abstract}

\section{Introduction}
Graph neural networks (GNNs), which are neural networks applied to graph tasks, have shown excellent performance in node classification, link prediction, and graph classification \cite{DBLP:journals/corr/abs-1901-00596}. However, the complexity of GNNs makes it difficult to clarify which structures of a given graph are crucial in GNN prediction. Thus, many methods have been proposed to interpret a trained GNN \cite{DBLP:journals/corr/abs-2012-15445}.

The widely used GNN explanation method is GNNExplainer \cite{DBLP:conf/nips/YingBYZL19}, which optimizes an edge mask such that the mutual information between the masked graph and the prediction is maximized when the size of the edge mask is constrained. PGExplainer \cite{DBLP:conf/nips/LuoCXYZC020} learns a neural network that outputs an edge mask under the condition that each edge parameter follows a Bernoulli distribution independently. SubgraphX \cite{DBLP:conf/icml/YuanYWLJ21} efficiently explores subgraphs using a Monte Carlo tree search to identify structures that play an important role in the prediction. These approaches can explain either or both important features and nodes. However, because these methods are based on non-convex optimization, a global optimal solution may not be obtained.

GraphLIME \cite{huang2020graphlime} is an extention of the novel LIME  method \cite{ribeiro2016should} for GNN, extracting the “dimensions of the node feature of a target node" using HSIC Lasso \cite{yamada2014high,climente2019block}, which is one of the supervised feature selection methods measuring the non-linear dependency between features and model outputs, in a locally interpretable manner. Because the HSIC lasso method is a convex method, it can obtain a global optimal solution. 
However, GraphLIME can only select important features and not interpret important structures. In GNNs, node interpretation is generally more important than explaining the important features. This is a major limitation of GraphLIME. 

In this study, we propose {\bf St}ructural{\bf GraphLIME}, which extends GraphLIME to handle graph-level tasks and select significant subgraphs. More specifically, we employ HSIC Lasso to solve the node selection problem using the perturbation of graphs. To extract subgraphs, we use the group and fused regularization. Finally, thanks to the HSIC Lasso formulation, the proposed method can also explain the graph series classification model. The key advantage of the proposed method is that it inherits the merits of GraphLIME and can be used for node and subgraph explanation tasks. Experiments demonstrate that the method can determine the significant structure of a given graph / graph series for a trained GNN.

\vspace{.1in}
\noindent {\bf Contribution}:
\begin{itemize}
    \item The proposed StGraphLIME extends GraphLIME to select important nodes and sub-graphs.
    \item A method for graph series classification is proposed as a novel task .
    \item Through experiments, it is demonstrated that StGraphLIME can successfully explain the node and sub-graphs for various synthetic and real-world tasks.
\end{itemize}

\section{Preliminary}
In this section, first the GNN explanation problem is formulated and then the GraphLIME method is introduced.

\subsection{Problem formulation}
For graph $G=(E,V,\boldX)$, $V$ denotes the set of nodes in $G$; $E$ denotes the set of edges in $G$; and $\boldX$ denotes the node feature matrix of $G$. The goal of this study is to propose explainable machine-learning models for the following GNN tasks.

\vspace{.1in}
\noindent {\bf GNN explanation for graph classification.} A GNN $\Phi$ in graph classification learns the parameters using a training dataset $\{(G_i, y_i)\}_{i=1}^N$, where $G_i=(V_i,E_i,\boldX_i)$ represents an instance graph, and $y_i$ represents the label of $G_i$, which are then used to estimate the label of an unseen graph. Given a trained GNN $\Phi$ and target graph $G=(V,E,\boldX)$ as an instance, the goal is to obtain a significant node set.

\vspace{.1in}
\noindent {\bf GNN explanation for graph series classification.}
A GNN $\Phi$ in graph series classification learns the parameters using a training dataset $\{(\{G_i^t\}_{t=1}^T, y_i)\}_{i=1}^N$, where $\{G_i^t\}_{t=1}^T$ represents an instance graph series of length $T$, and $y_i$ represents the label of $G_i$, which are then used to estimate the label of an unseen graph series. Given a trained GNN $\Phi$ and target graph series $\{G\}_{t=1}^T=\{(V^t,E^t,\boldX^t)\}_{i=1}^T$ as an instance, the goal is to obtain a significant pair of node sets and time points.

Note that finding significant node sets is a standard GNN explanation task, whereas graph series classification is a new GNN explanation task.

\subsection{GraphLIME}
GraphLIME is one of the methods used to interpret GNN prediction for node classification.

Let us denote $\boldX^{(v)} \in \mathbbR^{d \times n}$ as the feature vectors sampled to explain node $v$, and let $\boldy^{(v)} \in \mathbbR^n$ be the output of the GNN model $\Phi$. The sampled dataset can also be written as $\calD_v= \{(\boldx_i^{(v)}, y_i^{(v)})\}_{i=1}^n$, where $\boldX^{(v)} = [\boldx_1^{(v)}, \ldots, \boldx_n^{(v)}]$, and $\boldy^{(v)} = (y_1^{(v)}, y_2^{(v)}, \ldots, y_n^{(v)})^\top$. In the original paper, the $N$-hop network neighbors were considered to sample the neighboring nodes of a given node $v$. 

GraphLIME then selects important node features from $\calD_v$. More specifically, GraphLIME employs the HSIC lasso \cite{yamada2014high} to select important features \sloppy \cite{huang2020graphlime}, where the HSIC lasso was originally proposed to select important features of input $\boldx$ using a supervised dataset  \cite{yamada2014high}. The optimization problem of HSIC lasso is given as follows:
\begin{align*}
    \min_{\mathbf{\boldalpha}\in\mathbb{R}^{d}}&\hspace{.3cm}\biggl\{\frac{1}{2}\|\overline{\boldL}-\sum_{s = 1}^d\alpha_s\overline{\boldK}_s\|_{{\rm F}}^2 + \lambda\sum_{s=1} |\mathbf{\alpha}_s|\biggr\}, \\
    \text{s.t.}&\hspace{.3cm}\alpha_s \geq 0\ (\forall s),
\end{align*}
where $\lambda$ denotes a regularization parameter and $\overline{\boldL}=\boldH\boldL\boldH/{\|\boldH\boldL\boldH\|_{\rm F}}$ is the normalized Gram matrix of $\boldy$, $[\boldL]_{i,j} = s(y_i^{(v)}, y_j^{(v)})$ is a kernel for $\boldy^{(v)}$, \sloppy$ \overline{\boldK}_s=\boldH\boldK_s\boldH/\|\boldH\boldK_s\boldH\|_{\rm F}$ is the normalized Gram matrix of $s$-th feature, $[\boldK_s]_{i,j} = k(\boldx_i^{(s)}, \boldx_j^{(s)})$ is a kernel for the $s$-th feature vector $\boldx_i^{(s)}$, and $\boldH = \boldI_n - \frac{1}{n}\boldone_n \boldone_n^\top$ is the centering matrix.

In GraphLIME, a Gaussian kernel is employed for both the input and output kernels.
\begin{align*}
    k(\boldx_i^{(k)},\boldx_j^{(k)})&=\exp\left({-\frac{(\boldx_i^{(k)}-\boldx_j^{(k)})^2}{2\sigma_x^2}}\right),\\
    \ell(y_i,y_j)&=\exp\left({-\frac{\|y_i-y_j\|_2^2}{2\sigma_y^2}}\right),
\end{align*}
where $\sigma_x$ and $\sigma_y$ are Gaussian kernel widths. 

Because HSIC lasso is a convex method, a global optimal solution can be obtained. Thus, HSIC lasso can obtain stable results compared to non-convex and discrete optimization methods. Through experiments, it has been reported that GraphLIME outperforms GNNexplainer in feature-explanation tasks \cite{huang2020graphlime}. However, GraphLIME can only explain features and cannot be used to explain nodes. In this study, GraphLIME is extended to explain nodes.

\section{Proposed method}
In this section, we propose StGraphLIME, which is an extention of GraphLIME, to provide structural explanations for GNNs.

\begin{figure*}[tb]
    \centering
    \includegraphics[width=0.8\textwidth,keepaspectratio]{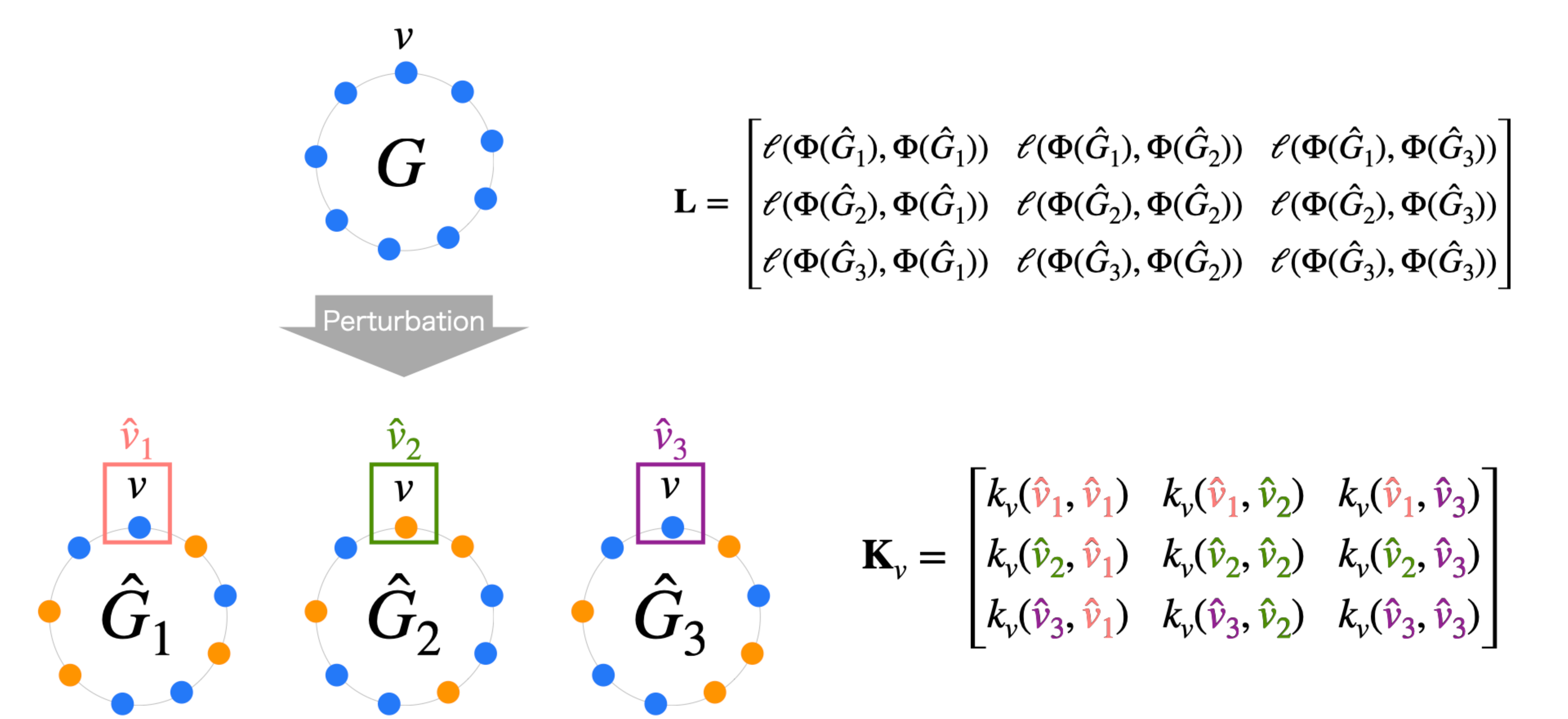}
    \caption{Calculating a gram matrix for a node and a prediction. The orange and blue nodes represent the perturbed node features.}
    \label{fig:gram}
\end{figure*}

\begin{table*}[t]
    \centering
        \caption{Examples of combinations of perturbation schemes and kernel functions. Various situations can be handled through appropriate selections.\label{tb:perturb}}
    \resizebox{\textwidth}{!}{
    \begin{tabular}{|c|c|c|c|}
     \hline
     \multicolumn{4}{|c|}{Variations of perturbation schemes and kernel functions} \\
     \hline
     Target & Perturbation & Features & Kernel. \\
     \hline
     The existence of certain nodes is important & Remove nodes & Binary features indicating whether a node exists & Delta kernel\\
     The existence of certain edges is important, & Remove edges & Binary features indicating whether the edge exists& Delta kernel\\
     The features of certain nodes are important & Add noise & Continuous node features & Gaussian kernel\\
     \hline
    \end{tabular}
    }
\end{table*}

\subsection{L1-regularized node explanation}
In this study, we propose a method to select graph features (e.g., nodes and edges) that are heavily dependent on the prediction of $\Phi$ using perturbation. Specifically, we first generate an auxiliary dataset $\{(\hat{G}_i,\Phi(\hat{G}_i))\}_{i=1}^M$ by adding perturbations to a target graph $G$, where $M$ is the number of perturbations. For perturbations, several nodes and edges can be removed from $G$ or noise can be added to node features. Figure \ref{fig:gram} shows an example of perturbation in graph $G$. In this example, three perturbed graphs ($M = 3$) are generated.

Then  HSIC lasso is used for the dataset to select nodes or edges that are significantly dependent on $\Phi(G)$. Now, we consider the detection of important nodes from $G$ is considered in graph classification. 
Let $\boldL$ and $\boldK_v$ be gram matrices for the prediction of $\Phi$ and node $v\in V$, respectively. Then, HSIC lasso for the explanations is formulated as follows:
\begin{align*}
    \min_{\boldalpha\in\mathbb{R}^{|V|}}&\hspace{.3cm}\biggl\{\frac{1}{2}\|\overline{\boldL}-\sum_{v\in V}\alpha_v\overline{\boldK}_v\|_{{\rm F}}^2 + \lambda\sum_{v\in V} |\mathbf{\alpha}_v|\biggr\}
    \\
    \text{s.t.}&\hspace{.3cm}\alpha_v \geq 0\ (\forall v\in V),
\end{align*}
where $\lambda$ denotes a regularization parameter and $\overline{\boldL}=\boldH\boldL\boldH/{\|\boldH\boldL\boldH\|_{\rm F}} \in \mathbbR^{M \times M},\ \overline{\boldK}_v=\boldH\boldK_i\boldH/\|\boldH\boldK_v\boldH\|_{\rm F} \in \mathbbR^{M \times M}$ are the normalized Gram matrices of perturbed samples.

Figure \ref{fig:gram} shows an example of building a gram matrix for target node $v$. Focusing on the significance of the existence of nodes for $\Phi$, $\boldK_v$ is calculated with the delta kernel from binary features, to determine the existing nodes in $\hat{G}_i$. Then, focusing on the significance of the continuous node features for $\Phi$, $\boldK_v$ is calculated using a gaussian kernel for the node features in $\hat{G}_i$. Table \ref{tb:perturb} lists the variations in perturbation schemes and kernel functions; $\boldL$ is calculated using predictive probability vectors (i.e., the output of $\Phi$) and the gaussian kernel. 

Note that we can also detect important edges in $G$ by replacing $\boldK_v$ with a gram matrix $\boldK_e$ associated with an edge $e \in E$. Thus, the proposed method is versatile in terms of explaining various GNN tasks. 

\vspace{.1in}
\noindent {\bf How to prepare an auxiliary dataset.}
Two strategies of adding perturbations to the original graph are proposed: (1) {\bf Random perturbation} and (2) {\bf Walk-based perturbation}. Random perturbations remove nodes or edges {\bf randomly} from the original graph or {\bf randomly} add noise to the node features of the selected nodes. Note that if the nodes are removed from a graph, the edges connected to the nodes are also deleted. On the other hand, walk-based perturbations add these perturbations to nodes or edges {\bf only on a random walk}. This means that walk-based perturbations make it possible to identify the functionality of the components in the original graphs.

\subsection{Group-regularized node explanation}
The L1 regularization method selects important node features independently. In this section, we show how to incorporate structural information into GraphLIME.

Group regularizations in lasso-like formulations enable the selection of input features in group units \cite{DBLP:conf/icml/JacobOV09,DBLP:journals/corr/abs-1110-0413}. In this study, group regularizations in HSIC lasso are used to identify crucial substructures of a graph for GNN prediction. HSIC lasso with group regularization for GNN explanations is formulated as follows:
\begin{align*}
   \min_{\boldalpha\in\mathbb{R}^{|V|}}&\hspace{.3cm}\biggl\{\frac{1}{2}\|\overline{\boldL}-\sum_{v\in V}\alpha_v\overline{\boldK}_v\|_{{\rm F}}^2+\lambda\sum_{\pi\in \Pi} \|\mathbf{\gamma}_\pi\|_2\biggr\},\\
   \text{s.t.}&\hspace{.3cm}\alpha_v \geq 0\ (\forall v\in V),
\end{align*}
where $\pi\subseteq V$ denotes an overlapping group composed of nodes and edges; $\mathbf{\gamma}_\pi$ denotes a vector composed of $\alpha_v$ corresponding to a node $v\in V$ contained in a group $\pi$; and $\lambda\geq 0$ denotes a regularization parameter. The edge version is obtained by replacing node $v$ with edge $e$. This objective can be solved in the same manner as latent group lasso \cite{DBLP:journals/corr/abs-1110-0413}. The effects of group regularizations on the latent group lasso are  discussed below.

\vspace{.1in}
\noindent {\bf The role of latent group reguralization.}
One of the crucial benefits of HSIC lasso in the context of feature selection is that interdependent features, in which the value of the HSIC is high, are less likely to be included in the selected features. This effect of latent group regularization in the context of GNN explanation prevents both overlapping groups from being selected as an explanation. 

\vspace{.1in}
\noindent {\bf How to construct groups.}
A walk-based strategy is also proposed to obtain groups for reguralization. In walk-base perturbations, nodes or edges on a random walk are used as a group. In addition, purposeful grouping is designed manually using prior knowledge of the target graph. For example, if the target graph is a molecular graph, it is possible to break it down into components based on the chemical functionalities.

\subsection{Fused-regularized node explanation}
A large target graph makes it difficult and costly to construct meaningful groups, both with simple random walks and prior knowledge. To mitigate this problem, another type of structural reguralization, the generalized fused lasso \cite{tibshirani2005sparsity} can be used as follows:
\begin{align*}
    &\min_{\boldalpha\in\mathbb{R}^{|V|}}\biggl\{\frac{1}{2}\|\overline{\boldL}\!-\!\sum_{v\in V}\alpha_v\overline{\boldK}_v\|_{{\rm F}}^2 + \lambda\sum_{v\in V} |\alpha_v|\\&+\mu\!\!\!\!\!\!\!\! 
 \sum_{(u_1,u_2)\in E}\!\!\!\!\!\!|\alpha_{u_1}\!-\!\alpha_{u_2}|\biggr\},
    \text{s.t.}~~\alpha_v \geq 0\ (\forall v\in V),
\end{align*}
where $\lambda $ and $ \mu\geq 0$ represent reguralization parameters. This formulation can also be extended to the edge version in a trivial manner. HSIC lasso with a generalized fused reguralization term is able to automatically cluster important nodes or edges. Consequently, significant substructures of the target graph are obtained without any grouping in advance.

\section{Related work}
There are lots of methods for interpreting GNN prediction \cite{DBLP:journals/corr/abs-2012-15445}.
CAM and Grad-CAM \cite{DBLP:conf/cvpr/PopeKRMH19} get node-level explanations for Graph Convolutional Networks.
These methods promise efficient results, however they require access to parameters of the target GNN and limit the architecture of it.
Compared to these approaches, StGraphLIME works in model-agnostic manner that treats GNNs as black-box models, which means we are able to obtain explanations for GNNs without knowing the internal architecture and parameters through model-agnostic explainablity methods.
Model-agnostic explainability methods for GNNs also appear in much literature.
The following approaches that we introduce work in model-agnostic manner.
GNNExplainer \cite{DBLP:conf/nips/YingBYZL19} provides structural explanation for GNNs by optimizing edge masks so that there is mutual information between the predictive distribution and the masked graph. 
PGExplainer \cite{DBLP:conf/nips/LuoCXYZC020} also obtains an edge-based explanation for a target graph by optimizing a neural network that estimates the edge mask.
GNNExplainer and PGExplainer do not guarantee connectivity of the explanation, which might lead to non-understandable results. Furthermore, optimizations in edge masks work well when connections of nodes are truly important in GNN prediction, whereas the results tend to be incomprehensible when nodes are present.
PGMExplainer \cite{DBLP:conf/nips/VuT20} generates a probabilistic graphical model for GNN explanation with considering influences on the prediction when perturbing node features.
SubgraphX\cite{DBLP:conf/icml/YuanYWLJ21} finds significant subgraphs for GNNs by using the Monte Carlo tree search algorithm and the Shapley values from game theory. 
Although SubgraphX always generates a connected graph as a human-friendly explanation, it cannot simultaneously detect separate important nodes.
GraphSVX \cite{DBLP:conf/pkdd/DuvalM21} computes fair contributions of node and its features with the Shapley Values and mask perturbations.

XGNN \cite{DBLP:conf/kdd/YuanTHJ20} is different from other methods in that XGNN provides ``model-level" explanations for GNNs.
Model-level explanation means returning a general explanation for the behavior of the trained GNN that do not depend on each instance, while StGraphLIME and the methods that we presented above make ``instance-level" explanations that aim to interpret the prediction of the trained GNN for a single instance.

GraphLIME \cite{huang2020graphlime} is the most similar to our method. GraphLIME selects significant ”dimensions of node feature" using HSIC lasso \cite{yamada2014high,climente2019block} for a local interpretation, while the proposed method ``StGraphLIME" is capable of selecting ”sub-structures of a graph" making the method an extension of GraphLIME. 

There are few works that explain GNNs designed especially for dynamic graphs.
DGExplainer \cite{DBLP:journals/corr/abs-2207-11175} calculates the contributions of subgraphs based on the LRP algorithm.
Wenchong et al. proposed to extend PGMExplainer to dynamic graphs by applying it to each snapshot and finding out dominant Bayesian networks \cite{DBLP:conf/globecom/HeV0T22}.
There works need to access the hidden states of a trained model, while our proposed StGraphLIME is a fully black-box explanation method that can be applied to dynamic graphs.

\section{Experiment}
In this section, the power of the proposed method is evaluated using five types of experiments.
These experiments include cases where nodes / edges / node features are important for regular, chemical graph, and graph series classifications.

\subsection{Setup}

\vspace{.1in}
\noindent {\bf Baselines.} 
A random explainer (which places a random score on each node), GNNExplainer \cite{DBLP:conf/nips/YingBYZL19}, PGExpaliner \cite{DBLP:conf/nips/LuoCXYZC020}, and SubgraphX \cite{DBLP:conf/icml/YuanYWLJ21} were adopted as the baselines for graph classification. In the graph series classificaion setting, we used the random explainer and occlusion explainer, where each node score is the change in the predictive score when setting the feature on the node to $0$. For a fair comparison, the edge score was converted to a node score, where the maximum value of the edge importance adjacent to the node is the node importance. The learning rate of GNNExplainer and PGExplainer was set to $0.005$, and the number of training iterations was set to $1000$.

\vspace{.1in}
\noindent {\bf Implementations.} 
In the experiments, the GNN models and GNNExplainer were implemented with PyG \cite{Fey/Lenssen/2019} and Pytorch Geometric Temporal \cite{rozemberczki2021pytorch} was used for temporal GNN models.
 As HSIC lasso solvers, scikit-learn \cite{scikit-learn}, SPAMS \cite{DBLP:conf/nips/MairalJOB10,DBLP:conf/icml/JenattonMOB10}, and CVXPY \cite{diamond2016cvxpy,agrawal2018rewriting} were used, and the DIG library \cite{DBLP:journals/jmlr/LiuLWXYGYXZLYLF21} was adapted for use in PGExplainer and SubgraphX. The implementation of HSIC lasso was based on the GraphLIME implementation\footnote{https://github.com/WilliamCCHuang/GraphLIME}.
The regularization parameters $\lambda, \mu$ of the proposed method were determined using a grid search on the validation data. When multiple parameters had the same validation score, the parameter that returned the smallest graph was selected because the smaller the output is, the more easily humans understand.

\subsection{Case study}
\begin{figure*}[t]
  \centering
  \begin{subfigure}[t]{0.25\textwidth}
    \centering
    \includegraphics[width=0.99\textwidth]{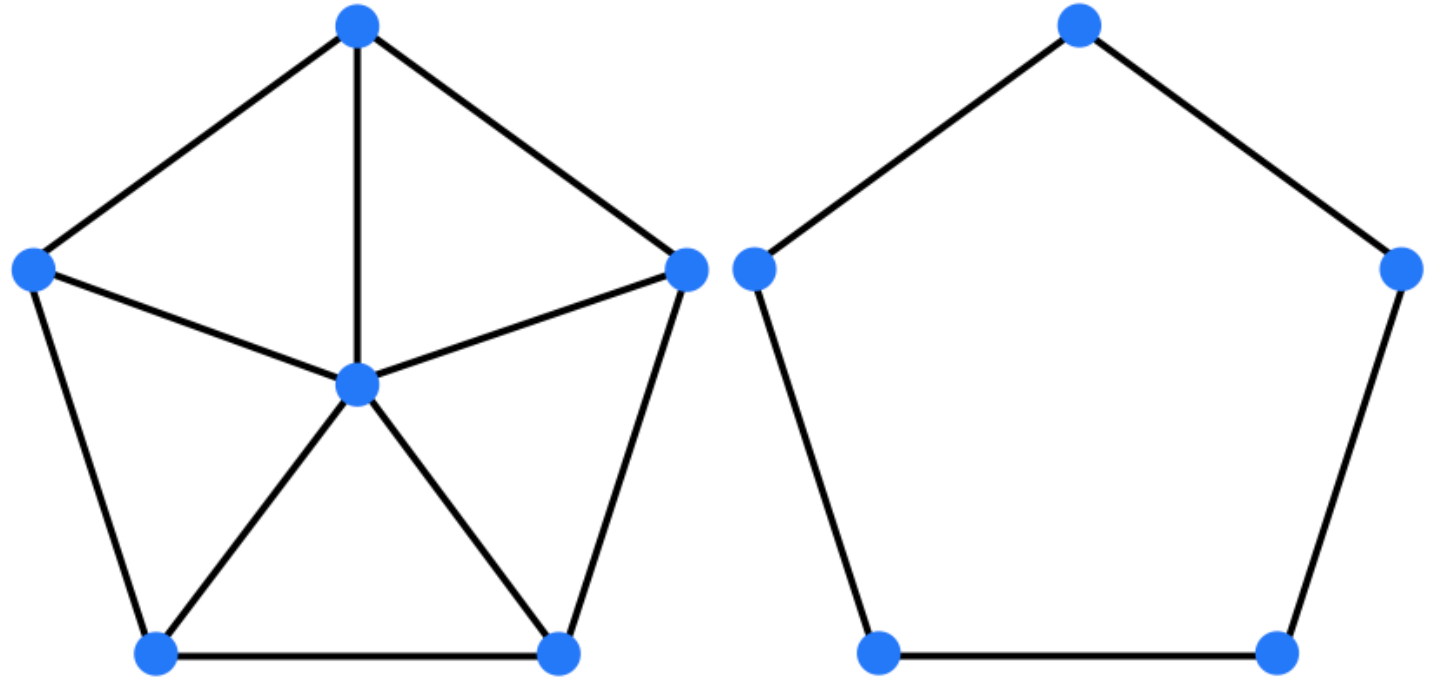}
    \caption{wheel graph (left) and cycle graph (right). }
  \end{subfigure}
  \begin{subfigure}[t]{0.25\textwidth}
    \centering
    \includegraphics[width=0.99\textwidth,keepaspectratio]{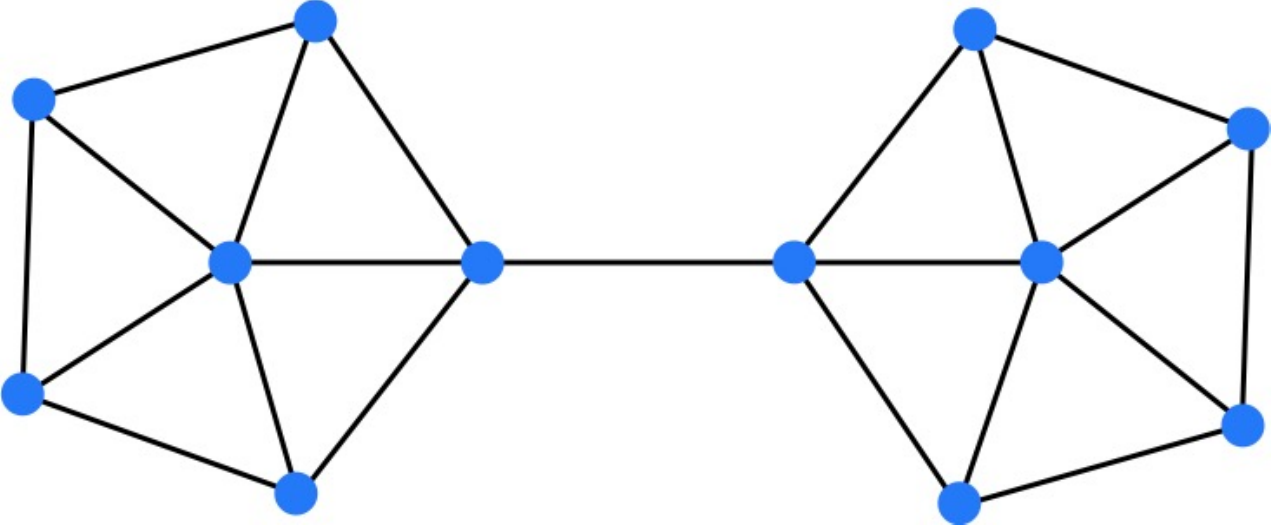}
    \caption{Two-connected-wheel graph.}
  \end{subfigure}
  \begin{subfigure}[t]{0.25\textwidth}
    \centering
    \includegraphics[width=0.99\textwidth,keepaspectratio]{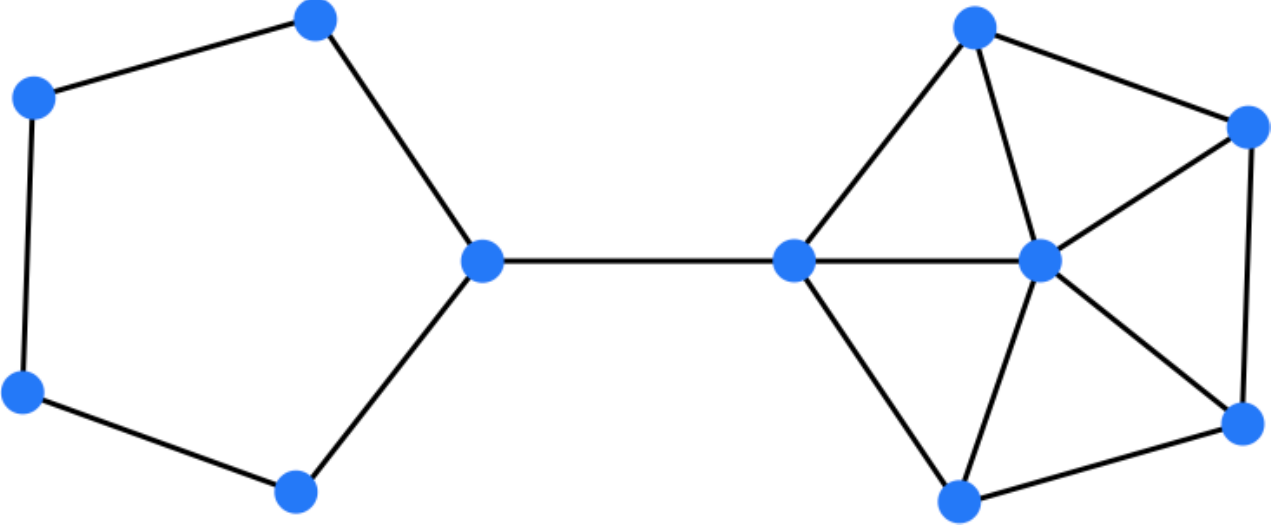}
    \caption{Mixed graph.}
  \end{subfigure}
      \begin{subfigure}[t]{0.25\textwidth}
    \centering
    \includegraphics[width=0.99\textwidth,keepaspectratio]{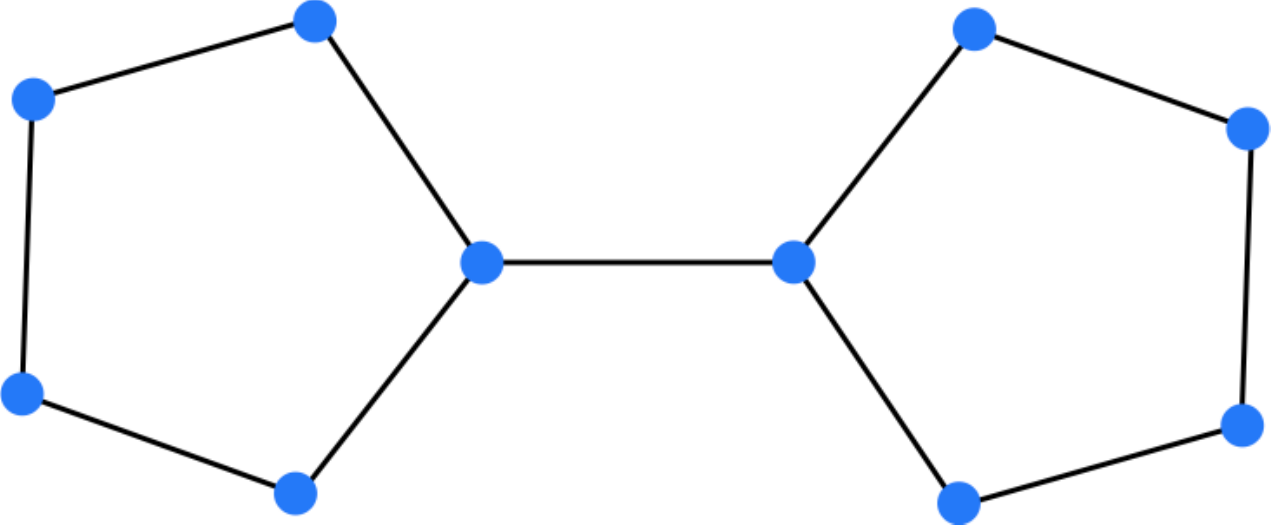}
    \caption{Two-connected-cycle graph.}
  \end{subfigure}
  \begin{subfigure}[t]{0.25\textwidth}
    \centering
    \includegraphics[width=0.99\textwidth,keepaspectratio]{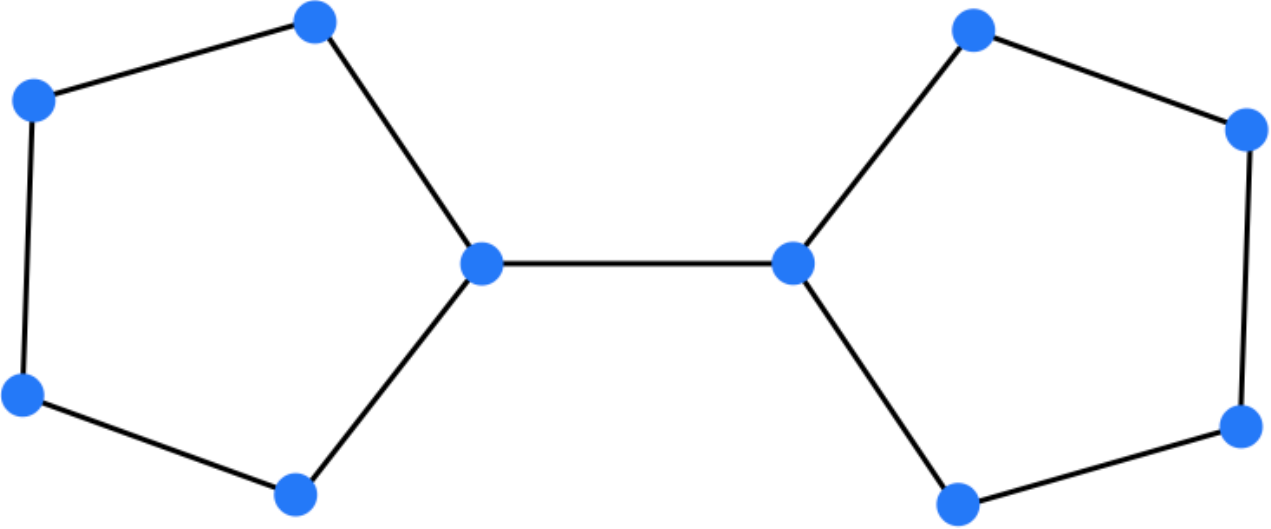}
    \caption{Glasses-like graph.}
  \end{subfigure}
      \begin{subfigure}[t]{0.25\textwidth}
    \centering
    \includegraphics[width=0.99\textwidth,keepaspectratio]{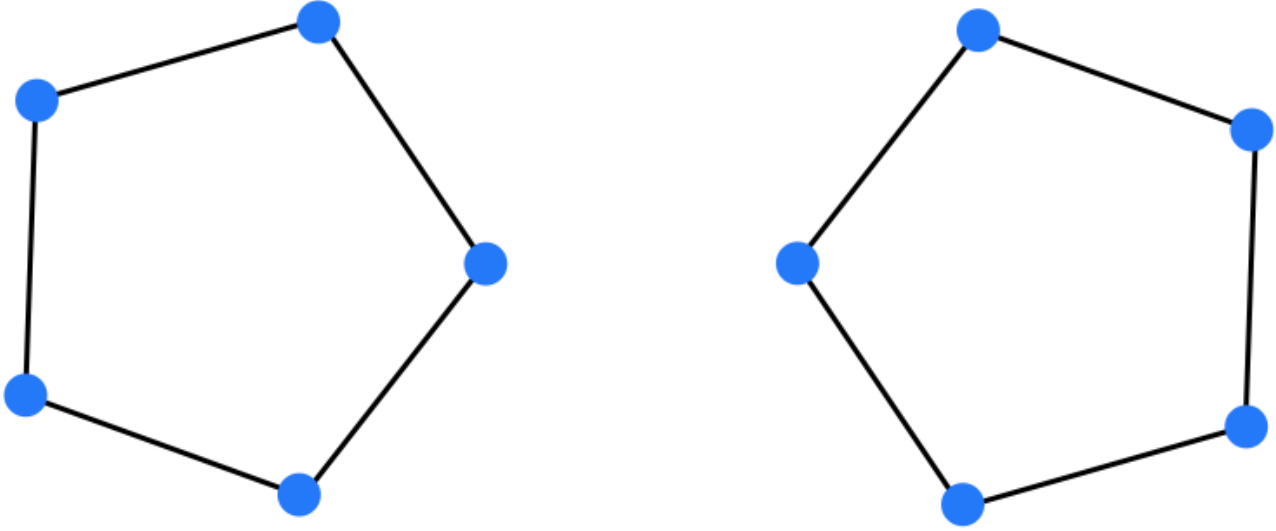}
    \caption{Two-disconnected-circles graph.}
  \end{subfigure}
\caption{An illustration of graphs. \label{fig:hub}}
 \end{figure*}

\vspace{.1in}
\noindent {\bf Case $1.1$: One node is target.} 
In case $1$, to confirm whether the proposed method can identify significant nodes, a trained GNN is used to classify the wheel and cycle graphs. Figure \ref{fig:hub} (a) shows an example of a cycle graph and wheel graph. The key difference between these two types of graphs is the presence of a ”hub" node. Explanation methods are expected to result in higher hub node scores than those of other nodes. Random perturbation was adopted by removing $2$ nodes, the delta kernel, and binary features with the $i$-th dimension indicating whether the $i$-th node is in $\hat{G}_i$ to compute $\mathbf{K}_v$ for the HSIC lasso. The size of the auxiliary dataset was $201$, and the original graph and predictive score were included. A GIN \cite{DBLP:conf/iclr/XuHLJ19} with three layers and max/mean/add global pooling was used as the model to be explained. The entire dataset consisted of cycle graphs with $5\sim 20$ nodes and wheel graphs with $6\sim 21$ nodes, which was split into training and testing data at a ratio of $9:1$. The trained GNNs classifed testing data with $100\%$ accuracy. The accuracy of detecting the true nodes, was evaluated using Top-$K$ Acc. which indicates whether the ground-truth node is among the top $K$ highest-scoring nodes. The wheel graphs with $6,7,8$ nodes were used as validation data, and the score was calculated from the rest. The grid search results are for $\lambda=10^{-6}$ in StGraphLIME (L1), $\lambda=10^{-2}$ in StGraphLIME (Group), and $\lambda=1.0, \mu=1.0$ in StGraphLIME (Fused). The means and standard deviations of the results were recorded for the three types of pooling layers. Table \ref{tb:hub} demonstrates that our method and SubgraphX can detect the hub nodes more accurately than the other methods.

\begin{table*}[tb]
\scriptsize
    \centering
    \caption{Results of the experiment for Case $1.1$ and $1.2$.}
    \label{tb:hub}
    \begin{tabular}{|c|c|c|c|c|c|c|}
     \hline
     & \multicolumn{2}{|c|}{Case $1.1$} & \multicolumn{4}{|c|}{Case $1.2$}\\
     \hline
     Methods & Top $1$ Acc. & Top $3$ Acc. & Top $2$ AAcc. & Top $6$ AAcc. & Top $2$ OAcc. & Top $6$ OAcc.\\
     \hline
     StGraphLIME (L1) & $0.98 (\pm 0.03)$ & $0.98 (\pm 0.03)$ & $0.81 (\pm 0.19)$ & $0.97 (\pm 0.05)$ & $0.92 (\pm 0.05)$ & $1.00 (\pm 0.00)$\\
     \hline
     StGraphLIME (Group) & $1.00 (\pm 0.00)$ & $1.00(\pm 0.00)$& $0.83 (\pm 0.05)$ & $0.88 (\pm 0.00)$ & $0.97 (\pm 0.05)$ & $1.00 (\pm 0.00)$\\
     \hline
     StGraphLIME (Fused) & $1.00 (\pm 0.00)$ & $1.00 (\pm 0.00)$& $0.75 (\pm 0.09)$ & $0.94 (\pm 0.05)$ & $0.92 (\pm 0.06)$ & $0.97 (\pm 0.00)$\\
     \hline
     SubgraphX & $1.00 (\pm 0.00)$ & $1.00 (\pm 0.00)$& $0.00 (\pm 0.00)$ & $0.56 (\pm 0.23)$ & $0.94 (\pm 0.03)$ & $1.00 (\pm 0.00)$\\
     \hline
     GNNExplainer & $0.02 (\pm 0.06)$ & $0.45 (\pm 0.13)$& $0.00 (\pm 0.00)$ & $0.11 (\pm 0.10)$ & $0.50 (\pm 0.29)$ & $0.94 (\pm 0.05)$\\
     \hline
     PGExplainer & $0.00 (\pm 0.00)$ & $0.00 (\pm 0.00)$& $0.00 (\pm 0.00)$ & $0.53 (\pm 0.16)$ & $0.56 (\pm 0.24)$ & $0.58 (\pm 0.19)$\\
     \hline
     Random & $0.02 (\pm 0.03)$ & $0.24 (\pm 0.03)$ & $0.00 (\pm 0.00)$ & $0.06 (\pm 0.05)$ & $0.11 (\pm 0.09)$ & $0.42 (\pm 0.10)$\\
     \hline
    \end{tabular}
\end{table*}

\vspace{.1in}
\noindent {\bf Case $1.2$: Two nodes are targets.} 
In this case, the explanation methods aim to detect the two important disconnected nodes. The synthetic dataset is composed of three types of graphs with binary labels. Figure \ref{fig:hub}(b)–(c) shows these $3$ patterns, two-connected-wheel graphs, two-connected-cycle graphs, and mixed graphs. The key difference between them is the presence of hub nodes in the cycle graphs on both sides. These graphs have labels according to whether there are hub nodes in each of them, which allows GNN to classify based on the existence of two hub nodes. Specifically, two-connected-wheel graphs and mixed graphs have positive labels, and two-connected-cycle graphs have negative labels. The explanation methods were evaluated using two metrics ($\rm AAcc.$ and $\rm OAcc.$) for the two-connected-wheel graphs; $\rm AAcc.$ indicates that the explanation method can identify both hub nodes, and $\rm {OAcc.}$ indicates that an explanation method can detect at least one of the two hub nodes. Random perturbation was adopted by removing $4$ nodes, the delta kernel, and binary features; the $i$-th dimension indicates whether the $i$-th node is in $\hat{G}_i$ to compute $\mathbf{K}_v$ for the HSIC lasso. The size of the auxiliary dataset was $201$, and the original graph and predictive score were included.  GIN \cite{DBLP:conf/iclr/XuHLJ19} with three layers and max/mean global pooling was used as the model to be explained. The reason for not adding pooling was because the GIN with added global pooling performed poorly. The number of nodes in a cycle graph, in a two-connected-cycle graph, and a mixed graph varied from $5$ to $15$, and the number of nodes in a wheel graph, in a two-connected-wheel graph, and a mixed graph varied from $6$ to $16$. The entire dataset was split into training and testing data at a ratio of $9:1$ and the trained GNNs classified the testing data with $100\%$ accuracy. To evaluate the accuracy of detecting the true nodes, Top-$K$ Acc. was used. The two-connected-wheel graph with $7,8$ nodes was used as validation data, and the score was calculated from the rest. The grid search results for $\lambda=10^{-8}$ in StGraphLIME (L1), $\lambda=10^{-7}$ in  StGraphLIME (Group), and $\lambda=1.0, \mu=10^{-7}$ in StGraphLIME (Fused) were calculated. The mean and standard deviation of the results were recorded for two runs each for two types of pooling layers. Table \ref{tb:hub} demonstrates that our methods could detect the two important hub nodes significantly better than the baselines. Although SubgraphX displayed a higher ability in identifying either hub node than other methods, the ability to identify both was inferior to our method and equivalent to GNNExplainer and PGExplainer. This is because SubgraphX always outputs a connected graph and is difficult to detect with a small number of candidates when the critical nodes are isolated.

\vspace{.1in}
\noindent {\bf Case $2$: One edge is target.} 
For case $2$, to confirm whether explanation methods can identify significant edges for prediction, a trained GNN was used to classify two-disconnected-cycle graphs and glass-like graphs. Figures \ref{fig:hub}(e) and (f) show an example of two-disconnected-circle graphs and a glass-like graph. The key difference between these graphs is the presence of a ”bridge" edge. Explanation methods are expected to result in higher scores on the bridge edge than on other nodes. A random explainer was adopted to remove $3$ edges, the delta kernel, and binary features with the $i$-th dimension indicating whether the $i$-th edge is in $\hat{G}_i$ to compute $\boldK_e$ for the HSIC lasso. The size of the auxiliary dataset was $201$, and the original graph and predictive score were included. A three-layer GIN \cite{DBLP:conf/iclr/XuHLJ19} with max global pooling was used as the model to be explained. The reason for not using additional pooling was because the GIN with added/mean global pooling did not show sufficient performance. The number of nodes in a cycle graph in a two-disconnected cycle graph and glass-like graph varied from $3$ to $10$. The entire dataset was split into training and testing data at a ratio of $9:1$ and the trained GNNs classified the testing data with $100\%$ accuracy. To evaluate the accuracy of detecting true nodes, Precision$@4$ was used. Glass-like graphs with $3,4$ nodes were used as validation data, and the score was calculated from the rest. The grid search results for $\lambda=10^{-7}$ in StGraphLIME (L1), $\lambda=10^{-2}$ in StGraphLIME (Group), and $\lambda=1.0, \mu=1.0$ in StGraphLIME (Fused) were calculated. The mean and standard deviation of the results of the three runs were recorded. Table \ref{tb:bridge} demonstrates that the proposed methods could detect the bridge better than baselines other than SubgraphX in Top-$5$ Acc.

\begin{table}[tb]
    \centering
        \caption{Results of the experiment for Case $2$ and $3$.}
    \label{tb:bridge}
    \resizebox{\columnwidth}{!}{
    \begin{tabular}{|c|c|c|c|}
     \hline
      & \multicolumn{2}{|c|}{Case $2$} & Case $3$\\ \hline
     Methods & Top $2$ Acc. & Top $5$ Acc. & Precision $@4$\\
     \hline
     StGraphLIME (L1) & $1.00 (\pm 0.00)$ & $1.00 (\pm 0.00)$ & $1.00 (\pm 0.00)$ \\
     \hline
     StGraphLIME (Group) & $1.00 (\pm 0.00)$ & $1.00 (\pm 0.00)$ & $0.98 (\pm 0.02)$\\
     \hline
     StGraphLIME (Fused) & $1.00 (\pm 0.00)$ & $1.00 (\pm 0.00)$ & $1.00 (\pm 0.00)$\\
     \hline
     SubgraphX & $0.39 (\pm 0.16)$ & $1.00 (\pm 0.00)$ & $0.96(\pm 0.03)$\\
     \hline
     GNNExplainer & $0.00 (\pm 0.00)$ & $0.11 (\pm 0.08)$ & $0.59(\pm 0.18)$\\
     \hline
     PGExplainer & $0.44 (\pm 0.21)$ & $0.44 (\pm 0.21)$ & $0.31 (\pm 0.09)$\\
     \hline
     Random & $0.00(\pm 0.00)$ & $0.11(\pm 0.16)$ & $0.24(\pm 0.06)$\\
     \hline
    \end{tabular}
    }
\end{table}

\begin{figure}[tb]
    \includegraphics[width=\columnwidth]{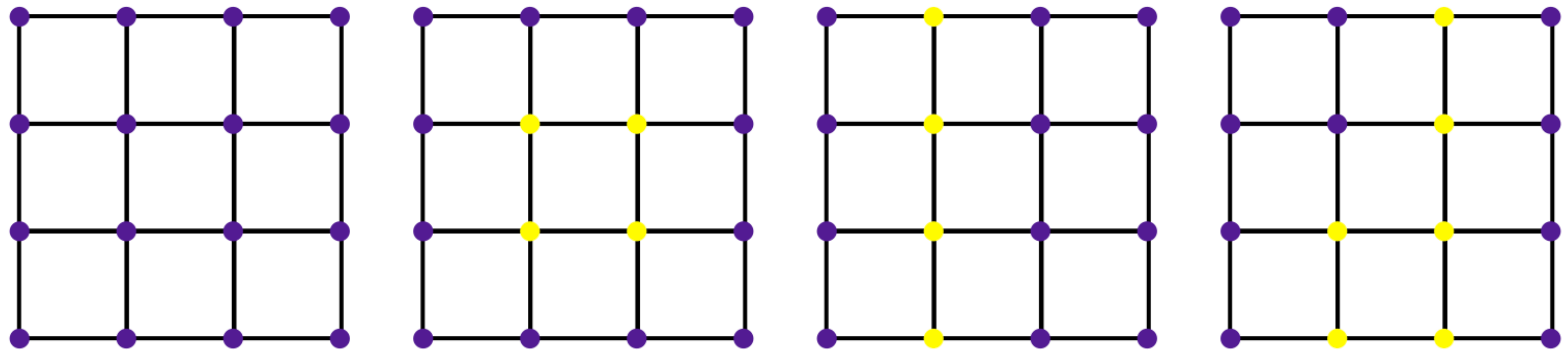}
    \caption{An illustration of none (left), rectangle (center left), line (center right), rectangle-line (right). Yellow nodes are nodes with $1$ features and purple nodes indicate nodes with $0$ features.}
    \label{fig:grid}
\end{figure}

\vspace{.1in}
\noindent {\bf Case $3$: Node features are target.} 
For Case $2$, the proposed methods were evaluated in a pattern recognition task on $4\times 4$ grid graphs. The four types of node feature patterns prepared were: none, rectangle, line, and rectangle-line, which is a combination of rectangles and lines. The four types of grid graphs are displayed in Figure \ref{fig:grid}. It is assumed that the important nodes for prediction are those with $1$ features. Rectangles and lines are targets for GNN explanation, and none and rectangle lines are used only in the training of leading nodes with $1$ features for prediction.
The random perturbation was adopted in the proposed methods by adding noise, a gaussian kernel, and perturbed node features. The size of the auxiliary dataset was $201$, and the original graph and predictive score were included. GIN \cite{DBLP:conf/iclr/XuHLJ19} with three layers and max/mean/add global pooling was used as the model to be explained. The size of the auxiliary dataset was $200$. The entire dataset was split into training and testing data at a ratio of $9:1$ and the trained GNNs classified the testing data with $100\%$ accuracy. Two rectangles and lines were used as validation data and the score was calculated from the rest. The grid search results in $\lambda=10^{-8}$ in StGraphLIME (L1), $\lambda=10^{-2}$ in StGraphLIME (Group), and $\lambda=1.0, \mu=1.0$ in StGraphLIME (Fused) were calculated.
The mean and standard deviation of the results were recorded for the three types of pooling layers. Table \ref{tb:bridge} demonstrates that the proposed methods can detect bridges more accurately than the other methods.

\vspace{.1in}
\noindent {\bf Case $4$: Chemical graph classification.} 
The MUTAG dataset\cite{debnath1991structure} is a chemical graph dataset for graph classification and is often used as a benchmark for GNN explanation tasks. The quantitative evaluation of GNN explanation methods is difficult for real-world graphs because the ground-truth explanations cannot be determined. Following \cite{DBLP:conf/icml/YuanYWLJ21}, the sparsity and fidelity scores were adapted to make quantitative evaluations. The fidelity for a set of pairs of graphs and an important node set $\{(G^i,M^i)\}_{i=1}^N$ is defined as follows:
\begin{align*}
    {\rm Fidelity}=\frac{1}{N}\sum_{i=1}^N\Phi(G^i)^*-\Phi(\hat{G^i}_{M^i})^*
\end{align*}
where $\Phi(G^i)^*$ denotes the prediction score of $\Phi$ for the $i$-th graph $G^i$, and $\hat{G^i}_{M^i}$ denotes a graph originating from $G^i$ with the feature of nodes in $M_i$ set to $0$. Fidelity is an indicator of the extent of worsening prediction when the node features are changed. The sparsity, for a set of pairs of graphs and an important node set $\{(G^i,M^i)\}_{i=1}^N$, is defined as follows:
\begin{align*}
    {\rm Sparsity}=\frac{1}{N}\sum_{i=1}^N1-\frac{|M^i|}{|G^i|},
\end{align*}
where $|G^i|$ denotes the number of nodes in $G^i$, Sparsity is an indicator of the ratio of the size of the selected important nodes $M^i$ to that of $G^i$. In general, fidelity and sparsity are trade-offs. The lower sparsity tends to increase the fidelity because more noise is added to the original graph and the prediction tends to worsen. This means that a comparison of the fidelity under the same sparsity values is necessary for a fair evaluation. On the other hand, because it is difficult to completely control sparsity, the sparsity-fidelity plot was evaluated according to Yuan et al. \cite{DBLP:conf/icml/YuanYWLJ21}. The average fidelity and sparsity were calculated in detail with the maximum number of important nodes to select, set to $10\%,20\%,30\%,40\%$, thereby plotting them as a sparsity-fidelity graph. Our methods, GNNExplainer, and PGExplainer selected up to the maximum number of specified nodes with positive importance scores in descending order of importance score. For SubgraphX, the maximum number of nodes in subgraph was set to the output of the aforementioned values.

\begin{figure*}[t]
  \centering
  \begin{subfigure}[t]{0.35\textwidth}
    \centering
    \includegraphics[width=0.99\textwidth,keepaspectratio]{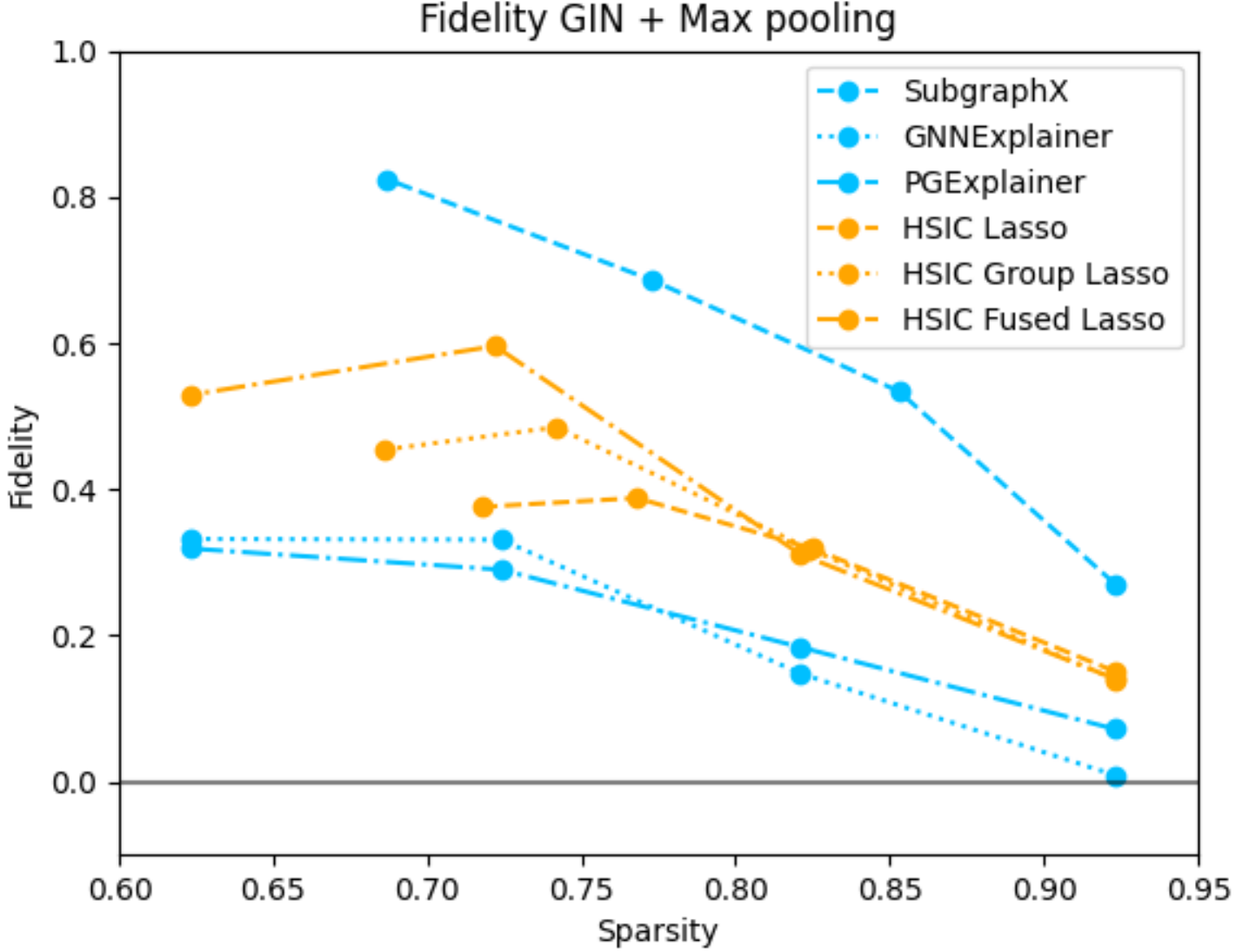}
    \caption{Fidelity (max-pooling). }
  \end{subfigure}
  \begin{subfigure}[t]{0.35\textwidth}
    \centering
    \includegraphics[width=0.99\textwidth,keepaspectratio]{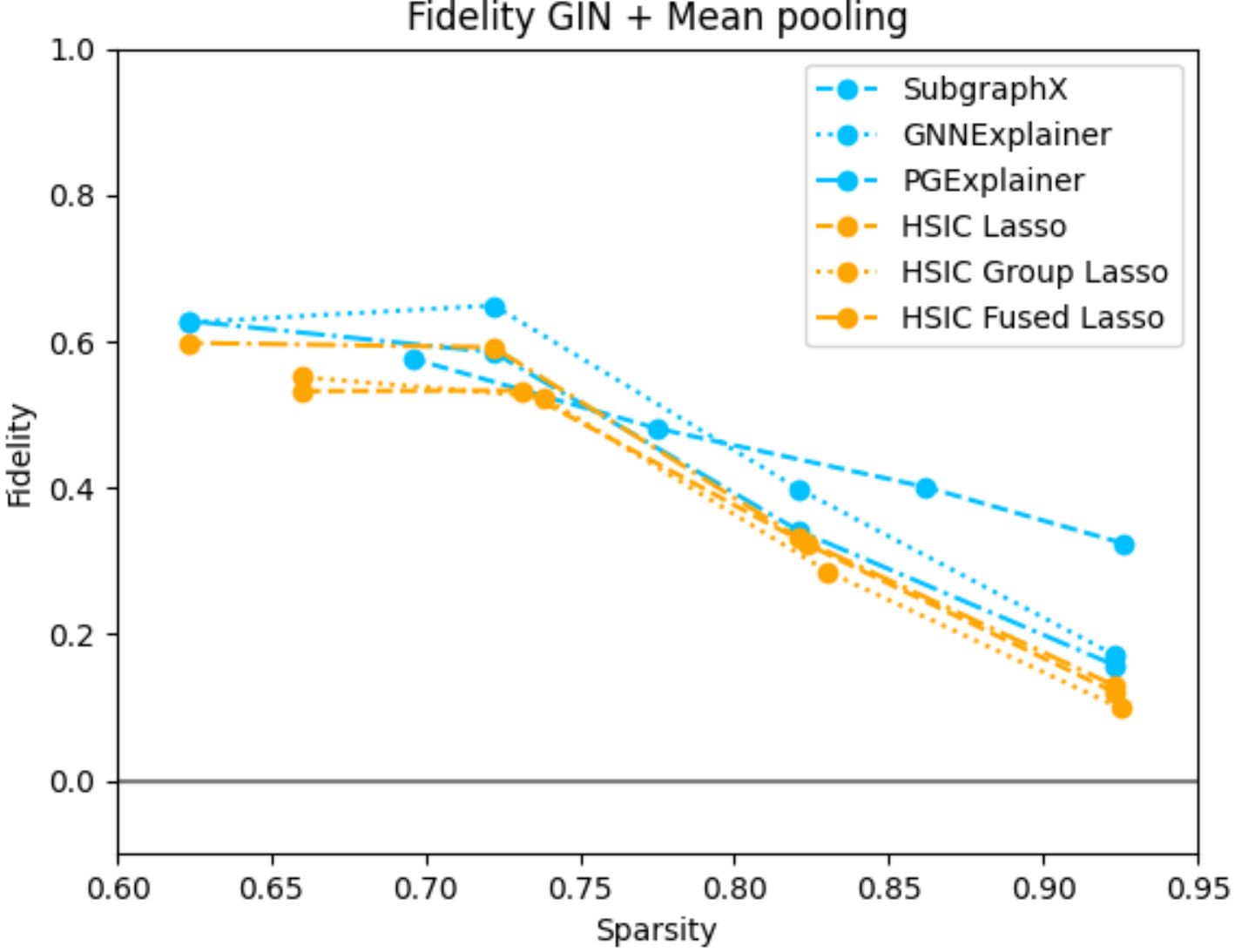}
    \caption{Fidelity (mean-pooling).}
  \end{subfigure}
  \caption{Quantitative results of the experiment for Case $4$.}
  \label{tb:fidelity}
\end{figure*}

Random-walk perturbation was adopted with the addition of noise, delta kernel, and categorical features with the $i$-th dimension indicating the $i$-th node feature to compute $\mathbf{K}_v$ for HSIC lasso. The size of the auxiliary dataset was $201$, and the original graph and predictive scores were included. GIN \cite{DBLP:conf/iclr/XuHLJ19} with three layers and max/mean global pooling was used as the model to be explained. Pooling was not added because the GIN with added global pooling performed poorly. The entire dataset was split into training and testing data at a ratio of $9:1$, and the trained GNNs classified the testing data with an accuracy of $ 89 \% $. A total of $10$ samples were used as validation data, and the score was calculated from the rest. The grid search results for $\lambda=10^{-9}$ in StGraphLIME (L1), $\lambda=10^{-5}$ in StGraphLIME (Group), and $\lambda=10^{-9}, \mu=10^{-2}$ in StGraphLIME (Fused) were calculated. Table \ref{tb:fidelity} lists the sparsity-fidelity plots for trained GNNs with max/mean pooling.
The proposed method outperformed GNNExplainer and PGExplainer, whereas SubgraphX displayed the best performance.

\begin{figure}[tb]
    \centering
    \includegraphics[width=0.99\columnwidth]{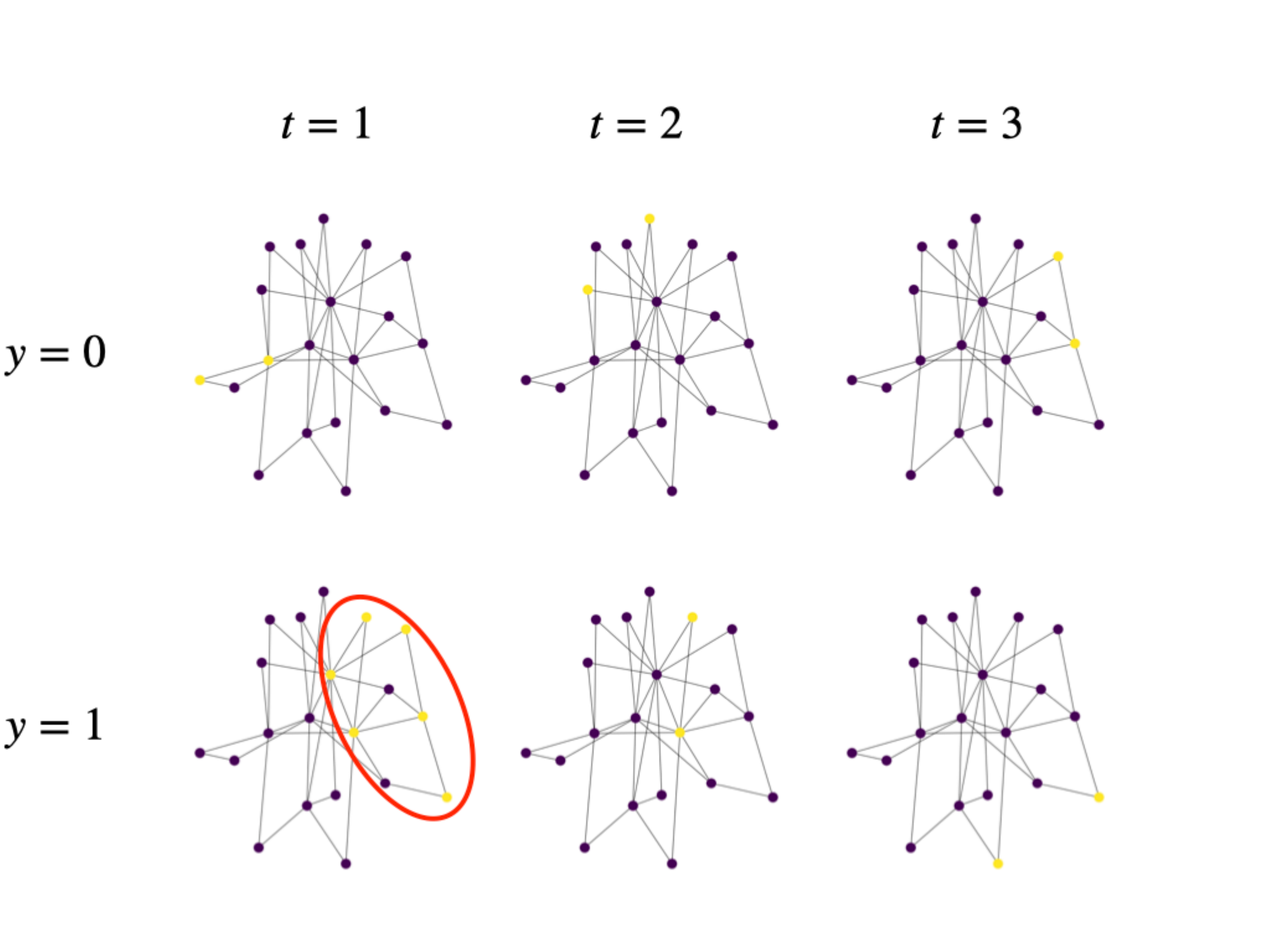}
    \caption{An illustration of graph series classification. The yellow nodes surrounded by the red line indicate that the graph at the bottom has a positive label.}
    \label{fig:series}
\end{figure}

\vspace{.1in}
\noindent {\bf Case $5$: Graph series classification.} 
The case $5$, was to confirm that the proposed method could identify nodes having significant features in sequential graph classification. These data comprise graph series of length $3$, where the structure of the graphs is fixed at all times for all series. The structure is determined using the Barab{\'a}si-Albert model \cite{barabasi1999emergence} with $20$ nodes. The number of series was $200$ and each node had a scalar feature of $0$ or $1$, which is considered a concrete feature.
If the features of the chunks of nodes at a certain time are $1$, the sample has a positive label. Otherwise, the label is negative. In other words, the key difference between positive and negative label graphs is whether there are node features on the chunk at a certain time in the graph series. Figure \ref{fig:series} shows an example of a graph series and its labels. Our method is expected to result in higher chunk scores than those of other nodes. The delta kernels were used and the node features were treated as labels to compute $\boldK_{v,t}$.

Random-walk perturbation was adopted in the proposed method by adding noise, a gaussian kernel, and perturbed node features. The size of the auxiliary dataset was $251$, and the original graph and predictive scores were included.
An $1$-layer T-GCN \cite{DBLP:journals/tits/ZhaoSZLWLDL20} with max/mean/add global pooling was used as the model to be explained. The entire dataset was split into training and testing data at a ratio of $9:1$ and the trained GNNs were classified as the testing data with $100\%$ accuracy. Another dataset was generated as validation data, and the score was calculated from the entire original dataset. The grid search results for $\lambda=10^{-6}$ in HSIC lasso, $\lambda=10^{-5}$ in HSIC group lasso, and $\lambda=1.0, \mu=10^{-1}$ in HSIC fused lasso were calculated. The means and standard deviations of the results were recorded for the three types of pooling layers. Table \ref{tb:series} demonstrates that our method can detect anomaly node groups more accurately than the other methods.

\begin{table*}[tb]
    \centering
    \scriptsize
    \caption{Results of the experiment for Case $5$. Precision$@TN$ represents precision score for nodes with top $TN$ scores, where $TN$ means the number of true nodes for each sample.}
    \label{tb:series}
    \begin{tabular}{|c|c|c|c|c|c|}
     \hline
     Methods & StGraphLIME(L1) & StGraphLIME(Group) & StGraphLIME(Fused) & Occulusion & Random\\ \hline 
     Precision$@TN$ & $0.77 (\pm 0.01)$ & $0.79 (\pm 0.01)$ & $0.80 (\pm 0.02)$ & $0.61 (\pm 0.07)$ & $0.12 (\pm 0.02)$\\
     \hline
    \end{tabular}
\end{table*}

\section{Discussion}

\vspace{.1in}
\noindent {\bf Strengths and weaknesses.}
The proposed methods work well in finding partial structural and feature differences compared with existing methods. This is because the existing methods measure the changes when perturbations are added to the original graph. On the other hand, the proposed methods work poorly for graphs without any “motif” which is the basis for GNN prediction. This issue is not only related to the proposed methods, but many other methods have the same problem.

\vspace{.1in}
\noindent {\bf Computational costs.}
The space complexity of the HSIC lasso is $O(n^2P)$, where $n$ denotes the number of nodes or edges in the target graph, and $P$ denotes the size of the auxiliary dataset. This cost can be reduced to $O(nBP)$ using block HSIC lasso \cite{climente2019block}, where $B\ll n$ denotes the block size. The power of detecting significant structures in the prediction strongly depends on $P$. More $n$ and less $P$ cause false positive results because of selection bias and decrease in the power of detecting significant structures due to lack of sufficient perturbation. For example, if a GNN classifies graphs based on a certain node set, to completely remove selection bias and maximize detection power, $P=O(2^n)$. If “the size of the target node is set to at most $K$", then $P={}_nC_K2^K=O(n^K2^K)$. If “the significant structure is a subgraph with $K$ nodes", then $P=(\#{ \rm subgraph \ with} \ K\ {\rm nodes})$. In the simplest case, if each node is perturbed at least once, then $P=O(n)$, and the total space complexity is cubic in this case.

\vspace{.1in}
\noindent {\bf Out-of-distribution problem.}
Structural perturbations in a graph based on the training data can generate out-of-distribution samples \cite{DBLP:journals/corr/abs-2010-13663}, causing misleading explanations. One way to mitigate this issue is to use feature perturbations instead of structural perturbations. In the chemical graph classification experiment, feature perturbation was adopted for this reason. Another possible approach is to calibrate the distribution used in HSIC lasso. Because our methods distributionally quantify the dependency between the graph structures and the prediction, correcting the distributions is a promising option.

\section{Conclusion}
A flexible method for the structure-based explanation of GNNs was proposed based on HSIC lasso. Group regularizations allow feature selection in the substructure units. The proposed methods perform better in cases where the key structural difference is evident, whereas a larger graph size makes it difficult for the proposed methods to detect the key differences.

\bibliography{arxiv}
\bibliographystyle{unsrt}

\end{document}